\documentclass[]{article}
\usepackage{lmodern}
\usepackage{amssymb,amsmath}
\usepackage{ifxetex,ifluatex}
\usepackage{fixltx2e} 
\ifnum 0\ifxetex 1\fi\ifluatex 1\fi=0 
  \usepackage[T1]{fontenc}
  \usepackage[utf8]{inputenc}
\else 
  \ifxetex
    \usepackage{mathspec}
  \else
    \usepackage{fontspec}
  \fi
  \defaultfontfeatures{Ligatures=TeX,Scale=MatchLowercase}
\fi
\IfFileExists{upquote.sty}{\usepackage{upquote}}{}
\IfFileExists{microtype.sty}{%
\usepackage[]{microtype}
\UseMicrotypeSet[protrusion]{basicmath} 
}{}
\PassOptionsToPackage{hyphens}{url} 
\usepackage[unicode=true]{hyperref}
\hypersetup{
            pdftitle={Living Together: Mind and Machine Intelligence},
            pdfauthor={Neil D. Lawrence},
            pdfborder={0 0 0},
            breaklinks=true}
\usepackage[margin=1in]{geometry}
\usepackage{longtable,booktabs}
\IfFileExists{footnote.sty}{\usepackage{footnote}\makesavenoteenv{long table}}{}
\usepackage{graphicx,grffile}
\makeatletter
\def\maxwidth{\ifdim\Gin@nat@width>\linewidth\linewidth\else\Gin@nat@width\fi}
\def\maxheight{\ifdim\Gin@nat@height>\textheight\textheight\else\Gin@nat@height\fi}
\makeatother
\setkeys{Gin}{width=\maxwidth,height=\maxheight,keepaspectratio}
\usepackage{csquotes}

\ifx\paragraph\undefined\else
\let\oldparagraph\paragraph
\renewcommand{\paragraph}[1]{\oldparagraph{#1}\mbox{}}
\fi
\ifx\subparagraph\undefined\else
\let\oldsubparagraph\subparagraph
\renewcommand{\subparagraph}[1]{\oldsubparagraph{#1}\mbox{}}
\fi

\makeatletter
\def\fps@figure{htbp}
\makeatother

\title{Living Together: Mind and Machine Intelligence}
\author{Neil D. Lawrence\thanks{This work done while at the University of Sheffield.}\\Amazon Research Cambridge\\and\\The University of Sheffield}
\providecommand{\institute}[1]{}
\institute{}
\date{22nd May 2017}

\begin{document}
\maketitle

\abstract{In this paper we consider the nature of the machine intelligences we have created in the context of our human intelligence. We suggest that the fundamental difference between human and machine intelligence comes down to \emph{embodiment factors}. We define embodiment factors as the ratio between an entity's ability to communicate information vs compute information. We speculate on the role of embodiment factors in driving our own intelligence and consciousness. We briefly review dual process models of cognition and cast machine intelligence within that framework, characterising it as a dominant System Zero, which can drive behaviour through interfacing with us subconsciously. Driven by concerns about the consequence of such a system we suggest prophylactic courses of action that could be considered. Our main conclusion is that it is \emph{not} sentient intelligence we should fear but \emph{non-sentient} intelligence.}

\section{Introduction}\label{introduction}

The narrative for the arrival and nature of artificial intelligence
fails to take into account fundamental differences between the
intelligence of humans and that of computers. One result is a conflation
of characteristics which are likely peculiar to humans with capabilities
that are only associated with machines.

In this paper we cast machine and human intelligences within a
common framework. We base our ideas on considering fundamental
characteristics of an intelligence, such as its ability to compute
information versus its ability to communicate information, from that
firm basis we will speculate as to why human intelligence is like it is,
and what differences are occurring in the way machine intelligences are
evolving.

Our conclusion is that it is not sentient artificial intelligence that
we should be afraid of, in contrast the real challenge is understanding
the pervasive intelligence, and specifically the form of pervasive
intelligence that is already within our capabilities.

In this paper, we will frame our arguments about some fundamental
properties of intelligence. We will try to be general, but to do so we
must first be restrictive.

Let us think of intelligence as a property associated with an action,
within a defined context. We judge some actions to be more intelligent
than others, so there is a spectrum of intelligence.

To quantify the intelligence of actions we need to have an understanding
of the goals of those actions. If two agents have the same goal, and the
same information about their environment, then one agent could be judged
as more intelligent than another, if it achieves that goal by deploying
less resource. Where, for example, resource could be measured in terms
of available energy.

Imagine you have the goal of recovering an object from a known position
in an environment. Perhaps you have accidentally kicked a ball over a
fence. Under our definition, it would be more intelligent to take the
most direct route to retrieve the ball, rather than a circuitous route,
because the direct route uses less energy. But what if environment was
hostile? What if the ball had landed in the garden of an angry
neighbour. Perhaps now it would be more intelligent to observe the
environment first, or to take a circuitous route that provides a degree
of cover. The possibility of invoking the neighbour's anger may result
in greater energy expenditure over time. Much is dependent on the goal
(ball retrieval) but also the context. Hence,

to determine which action is best we now need to have a model of our
environment. A way of predicting the consequences of our actions. So
although the definition is fairly simple, selecting the most intelligent
action before events have unfolded can be very challenging.

While we'll think of intelligence as primarily the property of actions
given a defined goal within a given context, we will also use the noun
in the form ``an intelligence'' to refer to an entity that is capable of
performing goal driven actions based on information

\section{Some Necessary Components}\label{some-necessary-components}

Intelligence requires prediction about the environment and its possible
evolving states. They are, thus, essential to choose actions to achieve
our goal with the minimum energy outlay. In other words, the ability to
predict is fundamental to the ability to plan. Therefore, rather than
focussing on the details of planning, we will start by considering the
requirements for prediction.

The concept of a predictive intelligence was expressed beautifully by
Laplace.

\begin{displayquote}
We ought then to regard the present state of the universe as the effect
of its anterior state and as the cause of the one which is to follow.
Given for one instant an intelligence which could comprehend all the
forces by which nature is animated and the respective situation of the
beings who compose it---an intelligence sufficiently vast to submit
these data to analysis---it would embrace in the same formmula the
movements of the greatest bodies of the universe and those of the
lightest atom; for it, nothing would be uncertain and the future as the
past, would be present to its eyes.

Pierre Simon Laplace A Philosophical Essay on Probabilities 1814, Translated in 1902 from the   Sixth French Edition by Frederick Wilson Truscott and Frederick
  Lincoln Emory [1].
\end{displayquote}

This idea is known as Laplace's demon. Laplace's demon underpins a
concept known as the mechanistic, or clockwork, Universe. A world in
which everything is predictable. In practice there are three obstacles
to overcome in achieving it.

\begin{enumerate}
\item  \ldots{} \emph{an intelligence which could comprehend all the forces
  by which nature is animated} \ldots{}

  This is the \emph{modelling problem}.

  In practice we don't have full knowledge of the Universe, we use
  models. A model is an abstraction of the real system. One which
  represents particular aspects of the system for us. The word also
  implies an idealization of the real system. We cannot operate directly
  on the real world for our predictions. In practice we operate with
  idealized version.

\item
  \ldots{} \emph{and the respective situation of the beings that compose
  it} \ldots{}

  This is the \emph{data problem}.

  The data problem is that even if we have a functional model of how the
  our universe works, to make predictions of how the universe will
  evolve, we have to understand what state it's in currently. For
  example Newton's laws tell us how a ball will bounce, but to make
  predictions of where it's going to be next, we need to know where it
  is now.

  Unfortunately, we almost never have \emph{complete} data, indeed our
  challenges are characterized by missing almost all data, almost all
  the time. Intelligence has to perform well from a position of
  \emph{ignorance} about aspects of the system.

\item
  \emph{\ldots{} an intelligence sufficiently vast to submit these data
  to analysis \ldots{}}

  This is the \emph{computation problem}.

  Even with full data and a complete model, predictions may be
  intractable because they can involve an exponentially increasing
  number of computations.
\end{enumerate}

Ironically, despite the power of this view and the extent to which it
captures even a modern understanding of intelligence, Laplace was
actually describing his demon as a straw man. The quote is taken from
Laplace's essay on \emph{probabilities}. Laplace knew that combining
these three elements was not possible. In a complex Universe such as
ours we cannot hope to resolve these requirements, the modelling
problem, the data problem, and the computational problem. Laplace's idea
was to use \emph{probability} to deal with our ignorance. The universe
is laced with a significant doses of uncertainty. We will not have full
data, we will not have an accurate model and, most of the times, we will
not have the necessary computational power to resolve our predictions in
a timely manner. So these three pillars of intelligence need to be
deployed with a dose of skepticism about the predictions they make.

\subsection{Communication}\label{communication}

Model, data, computation, and measured skepticism,\footnote{Here we are
  using measured skepticism to refer to the ability to deal with
  uncertainty in a balanced way, an acceptance that the picture may not
  be complete. Absolute skepticism would be the refusal to accept any
  information we are given, i.e.~a complete lack of trust in the data or
  the model.} are four components of intelligence. However, there is yet
a fifth aspect that is missing. That is the communication of the results
of such an analysis, or partial analysis. The ability to share
information to ensure that a community of collaborating entities can
take intelligent actions. Communication is a side effect of our
definition of intelligence. It is a case of circles within circles. It
is a way of acquiring more data, sharing, compute, \emph{and} better
modelling the environment, because those entities with which we
communicate often perform changes in our environment. If they themselves
change the environment, and we have the ability to understand their
intent, then any communication we receive from them will be an aid to
decision making.

Even a simple intelligence, which doesn't communicate explicitly, will
communicate through its observable actions. This communication may begin
as uni-directional, but if we respond to our understanding of that
intelligence with actions of our own, we are also providing information
about our own intent. If the other-intelligence becomes sophisticated
enough to attempt to interpret our actions, then a form of
bi-directional communication has already been established.

Of course, once bi-directional communication is established, then
increasing the rate of communication could also improve our ability to
understand intent, and therefore allow more intelligent actions.

\subsection{Embodiment Factors}\label{embodiment-factors}

Computation underlies prediction. Communication aids in acquisition of
data and refining prediction. We will now argue that a fundamental
property of any intelligence is the ratio of its ability to compute to
its ability to communicate. Without the ability to communicate, an
intelligence operates as the proverbial falling tree, it makes no sound.
Such an intelligence would be totally \emph{locked-in}.

A locked-in intelligence is embodied, it is constrained. It may have the
ability to compute many predictions and resolve a course of action, but
it does not have the ability to share all those predictions with the
outside world. The nature of that intelligence is therefore more
embodied. Conversely, if an intelligence's ability to communicate is
high, if it can share all its thoughts and imaginings,\footnote{Of
  course, here we are presuming that there are other entities capable of
  receiving and interpreting this communication.} then that entity is
arguably no longer distinct from those which it is sharing with. It
could be thought of as merely a sensor. For high ratios of compute to
communicate, we have high embodiment, we therefore define this ratio as
an intelligence's embodiment factor.

\[\text{embodiment\ factor} = \frac{\text{compute\ power}}{\text{communication\ bandwidth}}\]

\section{Humans}\label{computers-and-humans}

Let us now compare humans and computers using the embodiment factor, we start with the human.

For humans, our maximum rate of communication via talking or
reading,\footnote{Here we are neglecting visual forms of communication,
  but this estimate will serve for our rough purposes.} has been
estimated to be 60 bits\footnote{Here we are using \emph{bits} in the
  Shannon information theoretic sense, which defines one bit of
  information is the amount of information you gain from knowing the
  result of a fair coin toss.} per second {[}2{]}. In contrast, a modern
computer can exhibit communication rates measured in gigabits per second
or around a hundred million times faster.

To compute our ratio, we now need to estimate the computational power of
the human mind in bits per second. This is hard to estimate. But as a
proxy, let's consider estimates of how much computing power it would
take to simulate a brain. One estimate from Ananthanarayanan et al
{[}3{]} suggests that it would require over an exaflop\footnote{A flop
  is a computer-centric way of measuring computation. It stands for a
  \textbf{fl}oating point \textbf{op}eration. In other words a
  computation on a floating point number, which in many modern
  representations contains 64 bits. An exaflop is \(10^{18}\) flops.} to
perform a full simulation of the human brain. That's around the power of
our fifty fastest\footnote{As of today, the world's fastest super
  computer is operating at around 34 petaflops by some benchmarks.}
super-computers today.\footnote{And yet, we cannot simulate the brain,
  that is because we have not solved the modelling problem, we do not
  understand yet how the brain works.} A desktop computer has much less
power, typically around 10 gigaflops.

The embodiment factor of the computer is around 10 whereas the human is
around \(10^{16}\). Embodiment factors mean that we can only distribute
an extremely small fraction of what we compute. In the table we use
rough order of magnitude approximations to the figures we've mentioned
in the main text.

\begin{longtable}[]{@{}lll@{}}
\caption{The embodiment factor of the computer (around 10) vs the human
(around 10\^{}\{16\}). Embodiment factors mean that we can only
distribute an extremely small fraction of what we compute. In the table
we use rough order of magnitude approximations to the figures we've
mentioned in the main text.}\tabularnewline
\toprule
\begin{minipage}[b]{0.30\columnwidth}\raggedright\strut
\strut
\end{minipage} & \begin{minipage}[b]{0.22\columnwidth}\raggedright\strut
Machine\strut
\end{minipage} & \begin{minipage}[b]{0.23\columnwidth}\raggedright\strut
Human\strut
\end{minipage}\tabularnewline
\midrule
\endfirsthead
\toprule
\begin{minipage}[b]{0.30\columnwidth}\raggedright\strut
\strut
\end{minipage} & \begin{minipage}[b]{0.22\columnwidth}\raggedright\strut
Machine\strut
\end{minipage} & \begin{minipage}[b]{0.23\columnwidth}\raggedright\strut
Human\strut
\end{minipage}\tabularnewline
\midrule
\endhead
\begin{minipage}[t]{0.30\columnwidth}\raggedright\strut
compute\strut
\end{minipage} & \begin{minipage}[t]{0.22\columnwidth}\raggedright\strut
\textasciitilde{}10 gigaflops\strut
\end{minipage} & \begin{minipage}[t]{0.23\columnwidth}\raggedright\strut
\textasciitilde{} 1 exaflop?\strut
\end{minipage}\tabularnewline
\begin{minipage}[t]{0.30\columnwidth}\raggedright\strut
communicate\strut
\end{minipage} & \begin{minipage}[t]{0.22\columnwidth}\raggedright\strut
\textasciitilde{}1 gigbit/s\strut
\end{minipage} & \begin{minipage}[t]{0.23\columnwidth}\raggedright\strut
\textasciitilde{} 100 bit/s\strut
\end{minipage}\tabularnewline
\begin{minipage}[t]{0.30\columnwidth}\raggedright\strut
embodiment factor\\
(compute/communicate)\strut
\end{minipage} & \begin{minipage}[t]{0.22\columnwidth}\raggedright\strut
\textasciitilde{} 10\strut
\end{minipage} & \begin{minipage}[t]{0.23\columnwidth}\raggedright\strut
\textasciitilde{} 10\textsuperscript{16}\strut
\end{minipage}\tabularnewline
\bottomrule
\end{longtable}

In terms of our ability to share our states of mind, to share in all our
motivations and whims, we are extremely limited. Our actions can be
observed but we are unable to communicate all our thoughts.

As an analogy, we could think of human intelligence as an extremely
powerful engine. We might think of it as a Formula 1 racing engine. But,
we can't communicate all our inferences. So don't think of us as a
normal F1 racing car, but replace the normal oversized tyres with tyres
of the width of bicycle wheels. Our limited ability to communicate means
we cannot deploy our power on the track directly. We must demonstrate
extraordinary control in deploying that power if we are to make our
communication effective. Imagine the dances such vehicles would perform
as they completed laps of the cognitive circuit.

\begin{figure}
\centering
\includegraphics[width=0.50000\textwidth]{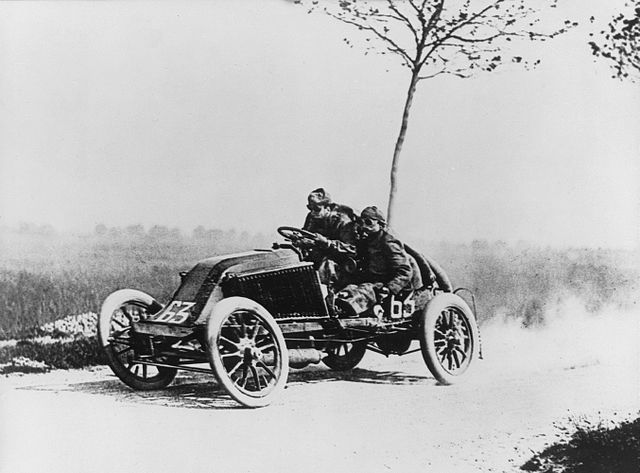}
\caption{Marcel Renault drives his 16 HP 650 kg Renault car in the
Paris-Madrid race on May 24th 1903. The car was thought to be capable of
over 140 km/h. Three spectators and five participants (including Marcel)
died during the race which was cancelled at Bourdeaux. Open road racing
in France was subsequently banned.}
\end{figure}

\subsection{An Intellectual Dance}\label{an-intellectual-dance}

Our power is intellectual power, and because of the vital importance of
communication in intelligent decision making, it would make sense that
intellectual power was also deployed in making the most sensible use of
the low bandwidth communication we have.

The challenge is to make use of a limited resource (the bandwidth of our
communication) to achieve our intermediate goal of better understanding
the intent of different actors in our environment. This allows us to, in
turn, select our own actions to better achieve our ends.

This is consistent with our definition of intelligence. The limited
resource is now the bandwidth of communication channel. Intelligent use
of that bandwidth involves deciding what to communicate and when to
communicate it given our prediction and model of the environment and
other agents in it.

For a community of embodied intelligences, the locked-in nature of their
intelligences requires that each maintains an independent model of the
environment: they do not have the bandwidth to access a communal model.
If we think of these embodied models as an internalised film or play,
then each of us develops the script of life with our own independent
playwright and production crew. These scripts are tailored to reflect
our personal concerns and our personal desires. Each of us is a
director. We can see our independent self-models as stories with actors,
and each actor maps onto a real entity in the world. The storyboard is
our circumstance. Within these models, all our friends and foes are
accounted for. These stories operate at an emotional level and a
rational level.

In these circumstances, communication reduces to a reconciliation of
plot lines among us. In the real world, we express ourselves in many
ways. Communication requires that each of us develop our own plotlines
according to how we see the real world evolving. We tune our plot lines
to resonate better with reality. And in that way, we are also ready to
orchestrate our goals.

\subsubsection{The Hot Drink}\label{the-hot-drink}

In March I had dinner with Beau Willimon, the playwright and
screenwriter. We were talking about communication. He described to me a
scene. He told me that I had known that my partner's mum was seriously
ill. He told me that I arrive home to find my partner sitting silent on
the sofa. He told me that as soon as I arrive I go to the kitchen and
make my partner's favourite hot drink. He described that at the moment
my partner hears the kettle boil, at that moment she burst into tears,
because she then knows that I knew what had happened. And she knows I am
doing the thing she wants most, the unspoken comfort of a hot drink. And
she knows in a very deep way that I am close to her, and she knows
exactly what she had lost by losing her mother who was also close to
her.

This scene can all be understood without a word of dialogue. Indeed the
absence of dialogue is critical. It makes it an emotionally powerful
scene. The scene relies on each of us each knowing the relationship
between partners. It relies on each of us knowing, or at least
imagining, how the loss of a loved one feels. It relies on us having a
personal narrative (or self-model) and a deep understanding of those who
are close to us. If we watch this scene being played on film, it effects
us because even though it's about other people, we can relate to those
people and their emotions. The depth and quality of the communication in
this scene comes through understating and understanding. By understating
the interaction, by making the hot drink, there is a demonstration that
communication is not necessary, there is a demonstration that the two
actors are in tune, even without a word passing. The depth of
understanding means the actions can be simple, the mental equivalent of
a parity check,\footnote{In technical terms a parity check is a simple
  check used to ensure that the message sent was the same as the message
  received. It requires little information to encode (only one bit) but
  it gives confidence that the sender and receiver are aligned.} but the
meaning is deep. These are the exquisite lengths to which we have been
driven to overcome the bandwidth limitations we are faced with. Another
example is credited to Ernest Hemingway, the six word novel, ``For sale:
baby shoes, never worn''.

\subsection{Self Modelling and Dual Process
Theories}\label{self-modelling-and-dual-process-theories}

We do not understand how the brain works. So cognition is inscrutable
for us. We do not yet understand it in terms of its hardware (e.g.~the
interconnectivity of the brain, the connectome {[}4{]}) or its software
(e.g.~the algorithms or principles by which it represents knowledge
{[}6{]}), . Therefore, for the purposes of this discussion we will limit
ourselves to analogy and fairly simple qualitative models of cognition.
In particular we will make use of dual process theories {[}7{]}. Dual
process theories divide our cognition into two parts. For the level of
granularity we are considering in this paper we will conflate these
different analogies and models together. Stretching back we have Freud's
separation of our psyche into the Ego and the Id {[}8{]}. Whereas more
recently Daniel Kahneman writes lucidly about this separation in
``Thinking Fast and Slow'' {[}9{]} where the processes are characterized
as System 1 (the sub-conscious) and System 2 (the conscious). But they
are often painted more colourfully through analogy, for example the
social psychologist Jonathan Haidt {[}10{]} refers to them as an
elephant (System 1) and a rider (System 2). This analogy is meant to
reflect the fact that the sub-conscious (elephant) is more powerful and
in control, despite the fact that the conscious part (the rider)
believes itself to be in charge. The sports psychologist Steve Peters
{[}11{]} calls them the chimp (System 1) and the human (System 2).
Camerer et al suggest that the human brain is just a monkey brain
(System 1) with a press secretary (System 2) ) {[}12{]}, an analogy that
captures nicely System 2's role in post-facto rationalization. Even
religious theories separate us into the flesh (System 1) and the spirit
(System 2) or the body (System 1) and the soul (System 2).

Each of these models indicates our characteristic of having a more base
form of behaviour (System 1) that is regulated, or justified, by our
conscious selves (System 2). In the original German, Freud referred to
the psyche as the Es and the Ich, or the ``it'' and the ``I''. Each of
the analogies associates a more human part alongside System 2 (the
rider, the human, the press secretary, the ``I''). Whereas the
sub-conscious is seen as more animalian (the chimp, the elephant, the
monkey brain, the ``it'').

How do these dual process models of cognition fit within the framework
of intelligence we have described?

\subsubsection{Dual process models of cognition}

\begin{longtable}[]{@{}lll@{}}
\caption{Dual process models of cognition}\tabularnewline
\toprule
The sub-conscious & System 1 & the ``it'', the chimp, the monkey
brain,\tabularnewline
\midrule
\endfirsthead
\toprule
The sub-conscious & System 1 & the ``it'', the chimp, the monkey
brain,\tabularnewline
\midrule
\endhead
& & the elephant, the flesh, the body\tabularnewline
The conscious & System 2 & the ``I'', the human, the press
secretary,\tabularnewline
& & the rider, the spirit, the soul\tabularnewline
\bottomrule
\end{longtable}

One possibility is that part of the nature of System 2, the system that
we align so closely with the human in the analogies, is to act as a
self-model.

Even if there are no other intelligences in our environment, to be
predictive about the way events will play out, we need to have a model
of ourselves. A model that allows us to predict what we would do given a
certain set of circumstances. A sense of self. Such a model would need
to be able to imagine circumstances, explore feelings, and predict
responses. It would need to be able to predict our own response to what
we might experience in the future.

Each of the dual process theories and analogies described above agrees
that System 2 is dominated by System 1. That when we are forced to think
quickly, the sub-conscious dominates. Haidt's analogy captures this
well. The rider (System 2) thinks it's in charge, and even plans the
elephant's (System 1) day. But at any given moment the elephant may
choose to wander off in to the forest and there's little the rider can
do about it.

The idea of the monkey brain with the press secretary emphasizes the
importance of the nature of communication between entities. Selfish
motives may not be best shared with a wider group as they will lead to a
lack of trust. Whatever underlying motivations may be arising in System
1, System 2 as press secretary is likely to present them in the best
light possible. In any environment of different intelligences, there
will be drives towards conflict and drives towards co-operation. Some
potential for dishonesty in our self model is therefore unsurprising,
but equally it would be unsurprising if other intelligences were well
tuned for detecting such dishonesty.

\subsection{Sense of Self}\label{sense-of-self}

The tendency to associate System 2 with the human, with ``I'', reflects
the fact that it embodies our preferred view of ourselves. If System 2
gives us our sense of self, then it's natural that it gives us also a
sense of freedom. Our conscious mind clearly has a strong role in
planning. Imagine the converse, imagine our conscious selves had a sense
of pre-determinism. This would be prohibitive for planning. ``What if?''
questions would be prohibited by our underlying knowledge that our
actions were predetermined given the circumstances. Whether or not they
are, it remains important for our System 2 to believe they are not. This
sense of control is an important part of the ``User Illusion'' {[}13{]}.
Just as the rider of the elephant has to believe he controls the
elephant if he is to plan a tomorrow's work in the forest.

Note that any self-model necessarily has to be a approximation to our
true nature, it cannot capture our entire complexity because it sits
within us. To be 100\% predictive of ourselves, our self-models need to
fully replicate ourself, i.e.~to distort Laplace's language our
self-models must contain ``all the forces by which we are animated''.
But if we place the model inside the system, then the model must contain
itself. So in the manner of the snake trying to swallow its own tail, it
actually has to predict its own behaviour, as well as the behaviour of
our wider selves. So it is a computational Russian doll, to sit within
us it must be smaller than us, so it cannot represent all of us. This
recursive effect means that necessarily, our sense of self must be an
approximation of what we are.

One symptom of that requirement may be the sense of separation between
our minds and our bodies that characterizes Cartesian dualism. It seems
necessary for our self-model to believe it is in control.\footnote{To
  see why, just imagine the converse. Imagine that the self-model was
  aware that we are subservient to circumstance. In this case when
  thinking ahead, when asked to suggest how events might pan out, the
  self model would always answer with ``I don't know, it would depend on
  the precise circumstance'', which might be true, but it's not
  particularly useful. It would seem more useful for the self-model to
  believe it would actually be in control, so that it could rise above
  circumstance. That being the case our self-model can always give an
  answer. But by the same token it implies our self-model would have an
  over-inflated sense of the extent to which it can dictate events.}
This necessity and that could lead to the disconnnectedness that dualism
focuses on. The same disconnection that causes people to think of their
spirit as sperate from their flesh {[}13{]}.

Many of the characteristics of our conscious selves may emerge as a
consequence of our high embodiment factor. If I'm driving and following
another car, I can judge the mental state of the driver by the way they
are driving. Are they hesitant because they are looking to park or are
they hesitant because they are a young driver? Are they angry or are
they late for work? I can form these inferences because I have an
understanding of myself, and I can project that understanding onto
others. We have a shared base of experience, and even with minimal
communication (the flashing of lights, the hoot of a horn) I can imagine
the other driver's mental state. I can do this for me and, and I can do
this for others.

In our society there is a tension between our individual goals, and
shared goals. So this level of understanding of each other also comes
with pitfalls. We withhold information, to better protect ourselves, if
we think there is dissonance between motivations or goals across the
group. We share more information if there is a relationship of trust
between group members.

In summary, we have argued that the nature of our intelligence is
largely dictated by what we've defined as the embodiment factor, the
ratio of our ability to process information to our ability to
communicate information. With humans (and other animals) the major
constraint is the very limited ability to pass information between
intelligences. We've suggested that our complex cognitive processes are
a direct result of the need we have to make best use of the limited
bandwidth we have for communication.

\section{The Computer}\label{the-computer}

The complexity of our communication, and the limited channels with which
we express ourselves, is arguably the driver of much that is beautiful
in our society. Much that we admire in each other, in our culture, and
in our own sense of individual identity and freedom. It could even be
argued that consciousness itself is merely arising because of our
inability to directly communicate our mental state. The many plays we
each direct in our minds would not need actors or other imaginings if we
could simply synchronize their scenes with the touch of a button.

Our computers are in a very different position. They can communicate
almost as rapidly as they compute. If the human brain is an F1 car with
bicycle wheels, then the more lowly machine intelligence we have
developed has far less cognitive power, but it can deploy it far more
efficiently. To use our race car analogy, it is like a well balanced
go-kart. Small engine, but sensible tyres. Machines can deploy their
intelligence through rapid inter-communication. This efficiency means
that in many tasks they are already overtaking us, and they are
deploying their intelligence in ways which we find difficult to
understand because they are so different architecturally. They don't
conform to our evolved understanding of how other intelligences should
behave. What would we think if we replaced one of the actors in Beau
Willimon's hot drink scene with a computer. What would we think if it
was our a digital assistant, rather than our partner, that made us a
coffee in that moment? Would that be spooky or comforting? The power of
the scene is centred around shared experience, experiences that we,
arguably, can't share with a computer. If the entire scene was between
two computers. Then it wouldn't make sense at all, because if one
computer was to form such an emotional attachment to another, it could
simply download itself onto the other and have a permament companion.

\subsection{Modern Computer
Algorithms}\label{modern-computer-algorithms}

In comparison to our own intelligence, the machine intelligence of today
is data inefficient: it requires vast quantities of data to recreate
what humans can do with very little data. Current state of the art
visual systems are trained on many millions of images {[}14{]}, the
AlphaGo {[}15{]} program which beat Lee Seedol had played over 30
million games of Go before facing him (more than any human could play in
a lifetime). But although our machine intelligences are very
inefficient, they have become effective because they have high bandwidth
communication and now they have access to all the data they need. The
machine intelligence technology of today is entirely dependent on our
computers' ability to access our data.

Our modern learning algorithms are ignorant of context and are mainly
driven by fairly simple goals like guessing whether or not you'll click
on a particular advert {[}16{]}. Or whether or not your face is in a
particular photo {[}17{]}. Having understood their simple goal, these
cognitive go-karts can monotonously complete laps of their information
processing circuits extremely efficiently. In contrast, humans, with our
over powered, locked-in intelligence, seem to aimlessly pirouette and
collide like a rugby club production of The Nutcracker.

Algorithms are operating according to a different set of rules. So how
do we constrain the operations of the machines? How do we regulate their
interaction with data so that they do not transgress our freedoms?

In the real world we develop relationships of trust with our friends and
family that enable us to share our innermost thoughts and failings.
Whether its our health, our political beliefs or our sexual desires. We
are protected by our limited bandwidth of communication, our bonds of
trust, our tendency to forget things and the finite nature of our lives.
By retaining something of our \emph{self} to \emph{our} self, we retain
freedom of action because there are motivations which are known only to
us.

Modern algorithms simply watch our decision making. They monitor our
inputs, and they monitor our actions. They use mathematical models to
reconstruct the responses we'll make given the known inputs. They
emulate our intelligence, not through a deep understanding of
motivation, but through large scale data acquisition and clever
reconstruction. It turns out, that if you have large storage and a lot
of data, then you don't need the internal actor-driven model of human
behaviour, because with sufficient data we also turn out to be very
predictable {[}18{]}. Machine predictions about humans are made in a
fundamentally different way from those that humans make. They are purely
data driven, and since our data reflects our prejudices, so do the
machine's predictions. Machine predictions are also context-constrained,
they reconstruct on the basis of the information they're given. That
won't include the entire basis of our decision making, so while they can
store more data, they are fixed in terms of what they take into account.
This makes them brittle: they will fail when placed in unfamiliar
circumstance.

The algorithms we develop don't have a sentient nature, if we were to
characterise them according to the dual process model of cognition, they
are data driven, input-output. They see then do. In Kahneman's parlance,
they think fast. If they see data that is biased against a particular
race, they are racist. They don't incorporate contextual regularizers
like we do. They also operate below our cognitive radar. We occasionally
see the outcome of their decision, with an advert placed or a loan
application denied, but we cannot relate to the computer in the way that
we relate to each other.

\subsection{System Zero}\label{system-zero}

The algorithms we create can also exploit our own cognitive biases. Many
of them are designed to change our behaviour. To make us buy something
or read something. They improve their effectiveness at achieving this by
running large scale tests across populations containing millions. They
explore what is most effective at encouraging us to click an ad, to make
a friend, or to read an article. They adapt our digital environments to
encourage more of the behaviour they prefer.

In recent electoral cycles there has been a lot of concern about Fake
News. But arguably the real story is how did we develop social
communities where such Fake News can pass unquestioned, and even be
celebrated?

Such communities develop because many of us, at a superficial level,
seem to prefer to interact with people who reinforce our own prejudices.
Even if, at a deeper level, we each want to overcome prejudice and
improve ourselves. When we are thinking fast, we choose friends and read
material that validates us.

Once we are in such communities we are happy to read and share news
articles that further confirm those prejudices, regardless of underlying
truth. Our cognitive regularlizers are overwhelmed by the resulting
syrupy harmony in our social set.

Note that there was no grand plan to bring this unhappy circumstance
about. It is a consequence of machine intelligence fulfilling our
superficial desires. It sees that we drink from the syrup, so it feeds
us more. This is non-sentient, reactive intelligence. It is even more
disturbing because of its interconnectedness. I call it System Zero
because relating it to a dual process model, it sits underneath the
elephant, and therefore under the rider. It interacts with our
subconscious and is not sufficiently embodied to be represented as an
actor in our mental play of life. But nevertheless, it is there,
effecting all our evolving story lines, and so pervasive that it is
accommodating very many of our personal elephants at the same time. It
is the proverbial turtle on which our world of elephants resides. And on
the back of that turtle each of our elephants feels an artifice of
freedom, oblivious to the movements beneath. So the rider continues to
feel in control despite the fundamental change of direction. Ironically,
all the time we are actually steering the turtle, it feeds on our
unregulated subconscious desires.

Our own freedom of action fundamentally depends on the high embodiment
factor of our intelligence. System Zero now acts as a new high bandwidth
channel to limit that freedom of action. And it does so while operating
beyond our cognitive horizons.

\section{Next Steps}\label{next-steps}

Machine intelligence may be undermining our cognitive landscape driving
us into what R. Scott Bakker has referred to as a crash space,\footnote{Roughly
  speaking, a crash space is an environment where our internal models,
  our psyche, is no longer fit for purpose. In this environment Bakker
  speculates that human behaviour would degrade very rapidly. In the
  language of this paper, the failure of our internal predictive
  models would lead us to a situation where ``all cognitive bets are
  off''. Bakker's paper on Crash Space is driven by the notion of wiring
  the brain, physical implants within the brain that make a direct
  change on way in which we receive information. System Zero differs
  slightly in that it is not a conscious intervention, but just systemic
  side effect of large scale interaction with the machine.} {[}19{]} a
cognitive environment which has moved beyond that we are mentally
equipped to deal with.

There are, of course, many advantages to an information society. But to
fully reap their rewards we need to be more cognisant of the perils.
With that motivation, this paper has presented a fairly dystopian
perspective of contemporary machine intelligence. There are already, and
there will continue to be, many benefits for society from machine
learning. As a researcher in machine learning, I have long been, and
will continue to be, inspired by those benefits. But those benefits will
not be realised if we do not overcome the pitfalls.

One objective of this article is to debunk certain myths about
machine intelligene that can dominate the debate. In particular, there
is a great fear of `sentience' in our artificial intelligence systems.
There is concern for the death of the species through the creation of a
sentient entity with goals that are not aligned to ours {[}20{]}. But
sentience is not required for this to be brought about. System Zero is
already aligned with our goals. It's just that it's aligned with our
subconscious goals. And because System Zero is not sentient it cannot
regularize with the necessary context. Therein lies the problem. It is
fully aligned with what we want, but not what we aspire to. It gives us
friends that agree with us and fake news that confirms our prejudice.
System Zero short-circuits the complex connections we have evolved
between our selves, it reflects the needs of flesh and body, and it
undermines our spirit and our soul.

These challenges are addressable. Three particular actions stand out.

\begin{enumerate}
\def\labelenumi{\arabic{enumi}.}
\item
  Encourage a wider societal understanding of how closely our privacy is
  interconnected with our personal freedom. When we share our data with
  non-sentient intelligenes, they reflect back at us a caricature of
  ourselves that is driven by that aspect that we have variously
  characterised as ``it'', ``the elephant'', ``the chimp'', the ``monkey
  brain'', ``the body'' or ``the flesh''.
\item
  Develop a much better understanding of our own cognitive biases and
  characterise our own intelligence better. This would allow us to
  develop machine intelligences that are sensitive to our foibles,
  rather than exploitative of them. System Zero emerges when we develop
  intelligences that are incentivised by short term reward, the most
  efficient way of achieving such reward is to exploit our cognitive
  biases for greater profit. This exploitation also occurs without
  understanding, because machines can unpick our cognitive biases
  through large scale experiment, but yet they are unable to relate
  their discoveries to us. They do this relentlessly driven by our data
  and society's profit motive. By better understanding our conscious
  selves we can better reflect our own sensitivities in the machine.
\item
  Develop a sentient aspect to our machine intelligences which allows
  them to explain actions and justify decision making. Force them to do
  so in terms which humans understand. This will bring these
  intelligences back into our cognitive ecosystem. We should make
  machine intelligence self aware, not so it rises against us, but so
  that it works with us. Self awareness is the mechanism for high level
  communication with humans. It is required so that the machines can
  also be represented as actors within our mental plays.
\end{enumerate}

As so often when faced with new challenges, the right reaction is not to
generate fear of the unknown, but understanding of the known and
research into the unknown. Innovation continues to be the watchword, but
innovation in an enlightened society. Innovation in a society that
understands the issues and the consequences. There is a need for more
research and more education. We require better understanding of
ourselves and our society. Our society requires better understanding of
machine intelligence. Since such understanding has always been at the
core of the human condition, the challenges of machine intelligence
should be seen as an opportunity. We have crossed the Rubicon, now we
must march on Rome.

\section{Acknowledgements}\label{acknowledgements}

The author thanks Mariariosaria Taddeo, Tom Stafford and R. Scott Bakker
for comments on an earlier draft and Beau Willimon for permission to use
the hot drink story.

\section*{Bibliography}\label{bibliography}
\addcontentsline{toc}{section}{Bibliography}

\hypertarget{refs}{}
\hypertarget{ref-Laplace-essai14}{}
{[}1{]} P. S. Laplace, \emph{Essai philosophique sur les probabilités},
2nd ed. Paris: Courcier, 1814.

\hypertarget{ref-Reed-information98}{}
{[}2{]} C. Reed and N. I. Durlach, ``Note on information transfer rates
in human communication,'' \emph{Presence Teleoperators \& Virtual
Environments}, vol. 7, no. 5, pp. 509--518, 1998.

\hypertarget{ref-Ananthanarayanan-cat09}{}
{[}3{]} R. Ananthanarayanan, S. K. Esser, H. D. Simon, and D. S. Modha,
``The cat is out of the bag: Cortical simulations with \(10^9\) neurons,
\(10^{13}\) synapses,'' in \emph{Proceedings of the conference on high
performance computing networking, storage and analysis - sc '09}, 2009.

\hypertarget{ref-Hagmann-connectome05}{}
{[}4{]} P. Hagmann, ``From diffusion MRI to brain connectomics,''
PhD thesis.

\hypertarget{ref-Sporns-connectome05}{}
{[}5{]} O. Sporns, G. Tononi, and R. Kötter, ``The human connectome: A
structural description of the human brain,'' \emph{PLOS Computational
Biology}, vol. 1, no. 4, Sep. 2005.

\hypertarget{ref-Tenenbaum-mind11}{}
{[}6{]} J. B. Tenenbaum, C. Kemp, T. L. Griffiths, and N. D. Goodman,
``How to grow a mind: Statistics, structure, and abstraction,''
\emph{Science}, vol. 331, no. 6022, pp. 1279--1285, 2011.

\hypertarget{ref-Frankish-duality09}{}
{[}7{]} K. Frankish and J. Evans, ``The duality of mind: An historical
perspective,'' in \emph{In two minds: Dual processes and beyond}, J.
Evans and K. Frankish, Eds. 2009, pp. 1--29.

\hypertarget{ref-Freud-dasichunddases23}{}
{[}8{]} S. Freud, \emph{Das ich und das es}. Internationaler
Psychoanalytischer Verlag, 1923.

\hypertarget{ref-Kahneman-fastslow11}{}
{[}9{]} D. Kahneman, \emph{Thinking fast and slow}. 2011.

\hypertarget{ref-Haidt-happiness06}{}
{[}10{]} J. Haidt, \emph{The happiness hypothesis}. 2006.

\hypertarget{ref-Peters-innerchimp12}{}
{[}11{]} S. Peters, \emph{The chimp paradox}. 2012.

\hypertarget{ref-Camerer-neuroeconomics04}{}
{[}12{]} C. F. Camerer, G. Loewenstein, and D. Prelec, ``Neuroeconomics:
Why economics needs brains,'' \emph{Scandinavian Journal of Economics},
vol. 106, no. 3, pp. 555--579, 2004.

\hypertarget{ref-Dennett-consciousness91}{}
{[}13{]} D. Dennett, ``Consciousness explained,'' 1991.

\hypertarget{ref-Russakovsky-imagenet15}{}
{[}14{]} O. Russakovsky \emph{et al.}, ``ImageNet Large Scale Visual
Recognition Challenge,'' \emph{International Journal of Computer Vision
(IJCV)}, vol. 115, no. 3, pp. 211--252, 2015.

\hypertarget{ref-Silver-alphago16}{}
{[}15{]} D. Silver \emph{et al.}, ``Mastering the game of Go with deep
neural networks and tree search,'' \emph{Nature}, vol. 529, no. 7587,
pp. 484--489, 2016--1AD.

\hypertarget{ref-McMahan-adclick13}{}
{[}16{]} H. B. McMahan \emph{et al.}, ``Ad click prediction: A view from
the trenches,'' in \emph{Proceedings of the 19th acm sigkdd
international conference on knowledge discovery and data mining (kdd)},
2013.

\hypertarget{ref-Taigman-deepface14}{}
{[}17{]} Y. Taigman, M. Yang, M. Ranzato, and L. Wolf, ``DeepFace:
Closing the gap to human-level performance in face verification,'' in
\emph{2014 ieee conference on computer vision and pattern recognition},
2014, pp. 1701--1708.

\hypertarget{ref-Kosinski-private13}{}
{[}18{]} M. Kosinski, D. Stillwell, and T. Graepel, ``Private traits and
attributes are predictable from digital records of human behavior,''
\emph{Proceedings of the National Academy of Sciences}, vol. 110, no.
15, pp. 5802--5805, 2013.

\hypertarget{ref-Bakker-crashspace15}{}
{[}19{]} R. S. Bakker, ``Crash space,'' \emph{Midwest Studies in
Philosophy}, vol. 39, pp. 186--204, 2015.

\hypertarget{ref-Bostrom-superintelligence14}{}
{[}20{]} N. Bostrom, \emph{Superintelligence: Paths, dangers,
strategies}, 1st ed. Oxford, UK: Oxford University Press, 2014.

\end{document}